\documentclass[11pt,a4paper]{article}
\usepackage[hyperref]{naaclhlt2018}
\usepackage{times}
\usepackage{latexsym}

\usepackage{url}

\aclfinalcopy % Uncomment this line for the final submission
\title{NAACL 2018 Tutorial -- The Interplay between Lexical Resources \\ and Natural Language Processing}

\author{Jose Camacho-Collados, Luis Espinosa-Anke\\
  School of Computer Science and Informatics \\
  Cardiff University \\
  {\tt camachocolladosj@cardiff.ac.uk} \\ 
  {\tt espinosa-ankel@cardiff.ac.uk} \\ \And
Mohammad Taher Pilehvar \\
  Language Technology Lab \\
  University of Cambridge \\
  {\tt mp792@cam.ac.uk} \\}

\date{}

\begin{document}
\maketitle
\begin{abstract}
  Incorporating linguistic, world and common sense knowledge into AI/NLP systems is currently an important research area, with several open problems and challenges. At the same time, processing and storing this knowledge in lexical resources is not a straightforward task. 
  This tutorial proposes to address these complementary goals from two methodological perspectives: the use of NLP methods to help the process of constructing and enriching lexical resources and the use of lexical resources for improving NLP applications. Two main types of audience can benefit from this tutorial: those working on language resources who are interested in becoming acquainted with automatic NLP techniques, with the end goal of speeding and/or easing up the process of resource curation; and on the other hand, researchers in NLP who would like to benefit from the knowledge of lexical resources to improve their systems and models. The slides of the tutorial are available at \url{https://bitbucket.org/luisespinosa/lr-nlp/}.

\end{abstract}

\urlstyle{tt}

\section{Description}

The manual construction of lexical resources is a prohibitively time-consuming process, and even in the most restricted knowledge domains and less-resourced languages, the use of language technologies to ease up this process is becoming a standard practice. NLP techniques can be effectively leveraged to reduce creation and maintenance efforts. In this tutorial we will present open problems and research challenges in these topics concerning the interplay between lexical resources and NLP. Additionally, we will summarize existing attempts in this direction, such as modeling linguistic phenomena like terminology, definitions and glosses, examples and relations, phraseological units, or clustering techniques for senses and topics, as well as the integration of resources of different nature. %The following topics are going to be covered in detail: terminology extraction, definition extraction, automatic extraction of examples, information extraction, hypernym discovery and taxonomy learning and topic/domain clustering. 

As far as the integration of lexical resources in NLP applications is concerned, we will explain some of the current challenges in Word Sense Disambiguation and Entity Linking, as key tasks in natural language understanding which also enable a direct integration of knowledge from lexical resources. We will explain some knowledge-based and supervised methods for these tasks which play a decisive role in connecting lexical resources and text data. Moreover, we will present the field of knowledge-based representations, in particular word sense embeddings, as flexible techniques which act as a bridge between lexical resources and applications. Finally, we will briefly present some recent work on the integration of this encoded knowledge from lexical resources into neural architectures for improving downstream NLP applications.%\footnote{The group at \url{goo.gl/JEazYH} is available as a discussion forum and place to receive the latest updates from this tutorial.}

\section{Outline}

\subsection{Introduction and Motivation}

Adding explicit knowledge into AI/NLP systems is currently an important challenge due to the gains that can be obtained in many downstream applications. At the same time, these resources can be further enriched and better exploited by making use of NLP techniques. In this context, the main motivation of this tutorial is to show how Natural Language Processing and Lexical Resources have interacted so far, and a view towards potential scenarios in the near future. 

As an introduction we first present an overview of current lexical resources, starting from the de facto standard lexical resource for English, i.e., WordNet \cite{Fellbaum:98}.
We provide a concise overview of WordNet, showing what synsets are and how the resource can be viewed as a semantic network.
We then briefly discuss some of the limitations of WordNet and discuss how these can be alleviated to some extent with the help of collaboratively-constructed resources, such as Freebase \cite{bollacker2008freebase}, Wikidata \cite{wikidata} and BabelNet \cite{NavigliPonzetto:12aij}.
As the main building block of these resources, we show how collaboratively-constructed projects, such as Wikipedia\footnote{\url{https://www.wikipedia.org/}} and Wiktionary\footnote{\url{https://www.wiktionary.org/}}, can serve as massive multilingual sources of lexical information.
The lexical resources session is concluded by a short introduction to the Paraphrase Database (PPDB) \cite{ganitkevitch2013ppdb,pavlick-EtAl:2015:ACL-IJCNLP3} and to a domain-specific lexical resource: SNOMED\footnote{\url{https://www.snomed.org/}}, which is one of the major ontologies for the medical domain.

The tutorial is then divided in two main blocks. First, we delve into NLP for Creation and Enrichment of Lexical Resources, where we address a range of NLP problems aimed specifically at improving repositories of linguistically expressible knowledge. Second, we cover different use cases in which Lexical Resources for NLP have been leveraged successfully. The last part of the tutorial focuses on lessons learned from work in which we tried to reconcile both worlds, as well as our own view towards what the future holds for knowledge-based approaches to NLP.

\subsection{NLP for Lexical Resources}

The application of language technologies to the automatic construction and extension of lexical resources has proven successful in that it has provided various tools for optimizing this often prohibitively costly and expensive process. NLP techniques provide end-to-end technologies that can tackle all challenges in the language resource creation and maintenance pipeline. In this tutorial we summarize existing efforts in this direction, including the extraction from text of linguistic phenomena like terminology, definitions and glosses, examples and relations, as well as clustering techniques for senses and topics. %We will additionally summarize recent work on the automatic integration of knowledge from heterogeneous resources such as BabelNet, ConceptNet, Uby or Yago.

\begin{enumerate}
    \item \textbf{Terminology extraction.} Measures for terminology extraction, the simple conventional tf-idf \cite{sparck1972statistical}, lexical specificity \cite{lafon:1980}, and more recent approaches exploiting linguistic knowledge \cite{hulth2003improved}.
    
    \item \textbf{Definition extraction.} Techniques for extracting definitional text snippets from corpora \cite{NavigliVelardi:10,boella2013extracting,Espinosa-Ankeetal2015weakly,Lietal:2016,anke2018syntactically}.
    
    \item \textbf{Automatic extraction of examples.} Description of example extraction techniques and designs on this direction, e.g.,  the GDEX criteria and their implementation \cite{kilgarriff2008gdex}.

    \item \textbf{Information extraction.} Recent approaches for extracting semantic relations from text: NELL \cite{carlson2010toward}, ReVerb \cite{fader2011identifying}, PATTY \cite{nakashole2012patty}, KB-Unify \cite{delli2015knowledge}.
    
    \item \textbf{Hypernym discovery and taxonomy learning.}  Insights from recent SemEval tasks \cite{Bordea2015SemEval2015T1,bordea2016semeval} and related efforts on the automatic extraction of hypernymy relations from text corpora \cite{velardi2013ontolearn,alfarone2015unsupervised,flati2016multiwibi,shwartz2016improving,EspinosaEMNLP2016,gupta2016revisiting}.
    
    \item \textbf{Topic/domain clustering techniques.} Relevant techniques for filtering general domain resources via topic grouping \cite{Roget:11,NavigliVelardi:04,babeldomains:2017}.
    
    \item \textbf{Alignment of lexical resources\footnote{Due to time constraints, items 7 and 8 were not presented during the tutorial.}.} Alignment of heterogeneous lexical resources contributing to the creation of large resources containing different sources of knowledge. We will present approaches for the construction of such resources, such as Yago \cite{Suchanek:07}, UBY \cite{gurevych2012uby}, BabelNet \cite{NavigliPonzetto:12aij} or ConceptNet \cite{speer2017conceptnet}, as well as other works attempting to improve the automatic procedures to align lexical resources \cite{MatuschekGurevych:2013,PilehvarNavigli:2014a}.

    \item \textbf{Ontology enrichment.} Enriching lexical ontologies with novel concepts or with additional relations \cite{jurgens2016semeval}.

\end{enumerate}

\subsection{Lexical Resources for NLP}
\label{sect:lex4nlp}

In addition to the (semi)automatic efforts for easing the task of constructing and enriching lexical resources presented in the previous section, we present NLP tasks in which lexical resources have shown an important contribution. Effectively leveraging linguistically expressible cues with their associated knowledge remains a difficult task. Knowledge may be extracted from roughly three types of resource \cite{Hovyetal:13}: unstructured, e.g. text corpora; semistructured, such as encyclopedic collaborative repositories like Wikipedia, or structured, which include lexicographic resources like WordNet. In this section we present some of the applications on which different kinds of lexical resource (including their combination) play an important role.

We begin this section by explaining some of the problems and challenges in Word Sense Disambiguation and Entity Linking, as key tasks in natural language understanding which enable the direct integration of knowledge from lexical resources. We describe the most relevant knowledge-based WSD systems, both based on definitions \cite{Lesk:86,BanerjeePedersen:03,basile-caputo-semeraro:2014:Coling} and graph-based \cite{agirre2014random,Moroetal:14tacl}; and supervised, both linear models \cite{ZhongNg:2010,iacobacci-pilehvar-navigli:2016:P16-1} and the most recent branch exploiting neural networks \cite{melamud2016context2vec,raganatoneural2017}. We present an analysis of the main advantages and limitations of each kind of approach \cite{raganato-camachocollados-navigli:2017:EACLlong}.

Then, we summarize the field of knowledge-based representations, in particular sense vectors and embeddings, as flexible techniques connecting lexical resources and downstream applications. We first present techniques which leverage WordNet as main source of knowledge \cite{chenunified:2014,RotheSchutze:2015,jauhar2015ontologically,johansson2015embedding,PilehvarCollier:2016emnlp} and also present other techniques exploiting multilingual resources such as Wikipedia or BabelNet \cite{iacobacci:2015,camacho2016nasari,mancini2017sw2v}.

Finally, we briefly present a few successful approaches integrating knowledge-based representations into downstream tasks such as sentiment analysis \cite{flekovasupersense}, lexical substitution \cite{cocos:sense2017} or visual object discovery \cite{young2017semantic}. As a case study, we present an analysis on the integration of knowledge-based embeddings into neural architectures via WSD for text classification \cite{pilehvaracl17}, discussing its potential and current open challenges.

%(Chen et al. 2014; Rothe and Schuetze, 2015; Camacho-Collados et al. 2016; Pilehvar and Collier, 2016; Mancini et al. 2017)

\subsection{Open problems and challenges}
\label{conclusion}

In this last section we introduce some of the open problems and challenges for automatizing the resource creation and enrichment process as well as for the integration of knowledge from lexical resources into NLP applications.

\section*{Instructors}

\paragraph{Jose Camacho Collados} is a Research Associate at Cardiff University. Previously he was a Google Doctoral Fellow and completed his PhD at Sapienza University of Rome. His research focuses on Natural Language Processing and, more specifically, on the area of lexical and distributional semantics. Jose has experience in utilizing lexical resources for NLP applications, while enriching and improving these resources by extracting and processing knowledge from textual data. On this area he has co-organized the SemEval 2018 shared task on Hypernym Discovery. Previously, he co-organized a workshop on “Sense, Concept and Entity Representations and their Applications” at EACL 2017 and a tutorial on the same topic at ACL 2016. His background education includes an Erasmus Mundus Master in Natural Language Processing and Human Language Technology and a 5-year BSc degree in Mathematics.

\paragraph{Luis Espinosa Anke} received his BA in English Philology in 2006 (Univ. of Alicante, Spain), and his PhD in Natural Language Processing in 2017 (Univ. Pompeu Fabra, Spain). He holds two MAs, one in English-Spanish Translation (Univ. of Alicante), and an Erasmus Mundus MA in Natural Language Processing (NLP) (Univ. of Wolverhampton and Univ. Autonoma de Barcelona). His research interests lie in the intersection between structured representations of knowledge and NLP, specifically computational lexicography and distributional semantics. He has co-organized the SemEval 2018 shared tasks on Hypernym Discovery and Multilingual Emoji Prediction. Previously, he co-organized the Spanish NLP conference (2014) and the Focused NER task (Open Knowledge Extraction challenge) at ESWC 2017.

\paragraph{Mohammad Taher Pilehvar} is a Research Associate at the University of Cambridge. Taher's research lies in lexical semantics, mainly focusing on semantic representation and similarity. In the past, he has co-instructed three tutorials on these topics (EMNLP 2015, ACL 2016, and EACL 2017) and co-organised three SemEval tasks. He has also co-authored several conference (including two ACL best paper nominations, at 2013 and 2017) and journal papers, including different semantic representation techniques based on heterogeneous lexical resources.

%\section*{Acknowledgments}

%The acknowledgments should go immediately before the references.  Do
%not number the acknowledgments section. Do not include this section
%when submitting your paper for review. \\

\bibliography{naaclhlt2018}
\bibliographystyle{acl_natbib}

%\appendix

%\section{Supplemental Material}
%\label{sec:supplemental}
%Submissions may include resources (software and/or data) used in in the work and described in the paper. Papers that are submitted with accompanying software and/or data may receive additional credit toward the overall evaluation score, and the potential impact of the software and data will be taken into account when making the acceptance/rejection decisions. Any accompanying software and/or data should include licenses and documentation of research review as appropriate.

%\section{Multiple Appendices}
%\dots can be gotten by using more than one section. We hope you won't
%need that.

\end{document}